\documentclass[journal]{IEEEtran}

\usepackage{colortbl,booktabs}
\usepackage{amsmath}
\usepackage{times}
\usepackage{epsfig}
\usepackage{graphicx}
\usepackage{amsmath}
\usepackage{amssymb}
\usepackage{color}
\usepackage{wrapfig}
\usepackage{multirow}
\usepackage{multicol}
\usepackage{cite}
\usepackage{algorithm}
\usepackage{algorithmic}
\usepackage{subfigure}
\usepackage[hyphens]{url}  

\ifCLASSINFOpdf
\else
\fi
\begin{document}

\title{Exposure Fusion for Hand-held Camera Inputs with Optical Flow and PatchMatch}

\author{Ru~Li,~\IEEEmembership{Member,~IEEE,}
        Guanghui~Liu,~\IEEEmembership{Senior Member,~IEEE,}
        Bing~Zeng,~\IEEEmembership{Fellow,~IEEE,}
        Shuaicheng~Liu,~\IEEEmembership{Member,~IEEE}
        
\thanks{
Ru~Li is with School of Computer Science and Technology, Harbin Institute of Technology, Weihai 264209, China.

Guanghui~Liu, Bing~Zeng and Shuaicheng~Liu are with School of Information and Communication Engineering, University of Electronic Science and Technology of China, Chengdu 611731, China.

}
}


\maketitle

\begin{abstract}
This paper proposes a hybrid synthesis method for multi-exposure image fusion taken by hand-held cameras. Motions either due to the shaky camera or caused by dynamic scenes should be compensated before any content fusion. Any misalignment can easily cause blurring/ghosting artifacts in the fused result.
Our hybrid method can deal with such motions and maintain the exposure information of each input effectively. In particular, the proposed method first applies optical flow for a coarse registration, which performs well with complex non-rigid motion but produces deformations at regions with missing correspondences. The absence of correspondences is due to the occlusions of scene parallax or the moving contents. To correct such error registration, we segment images into superpixels and identify problematic alignments based on each superpixel, which is further aligned by PatchMatch. The method combines the efficiency of optical flow and the accuracy of PatchMatch. After PatchMatch correction, we obtain a fully aligned image stack that facilitates a high-quality fusion that is free from blurring/ghosting artifacts.
We compare our method with existing fusion algorithms on various challenging examples, including the static/dynamic, the indoor/outdoor and the daytime/nighttime scenes. Experiment results demonstrate the effectiveness and robustness of our method.
\end{abstract}

\begin{IEEEkeywords}
Multi-exposure fusion, optical flow, patch match
\end{IEEEkeywords}

%
\IEEEpeerreviewmaketitle

\section{Introduction}\label{sec:introduction}

Dynamic ranges of natural scenes are much wider than those captured by commercial imaging products. Recent digital cameras often fail to capture the irradiance range that is visible to the human eyes. High dynamic range (HDR) imaging techniques have attracted considerable interest due to they can overcome such limitation. HDR imaging has been increasingly used in consumer electronics, industrial, security and military applications~\cite{darmont2012high}. Directly capturing and displaying an HDR image is expensive. A relatively cheap way is to capture a stack of different exposure images and then merge them together. There are two main categories to conduct the synthesis. One is to reconstruct an HDR image through the camera response function (CRF) and then apply the tone mapping for the display~\cite{Debevec1997Recovering,Reinhard2002Photographic,Reinhard2010High}.
This two-phase workflow has some advantages such as generating images with virtual exposure and desired appearance. However, the process is usually not as convenient as multi-exposure fusion (MEF). When estimating CRF, it is challenging to faithfully recover some particular parameters that can break the self-similar and exponential ambiguities~\cite{grossberg2003determining}. Sometimes, the process even requires a special hardware for the accurate
solution~\cite{zhao2015unbounded}. In contrast, the other category, MEF, can directly synthesize a low dynamic range (LDR) image from several different exposure images that are more informative and detailed than any input. Our method belongs to the second category.

\begin{figure}[t]
	\centering
	\includegraphics[width=1.0\linewidth]{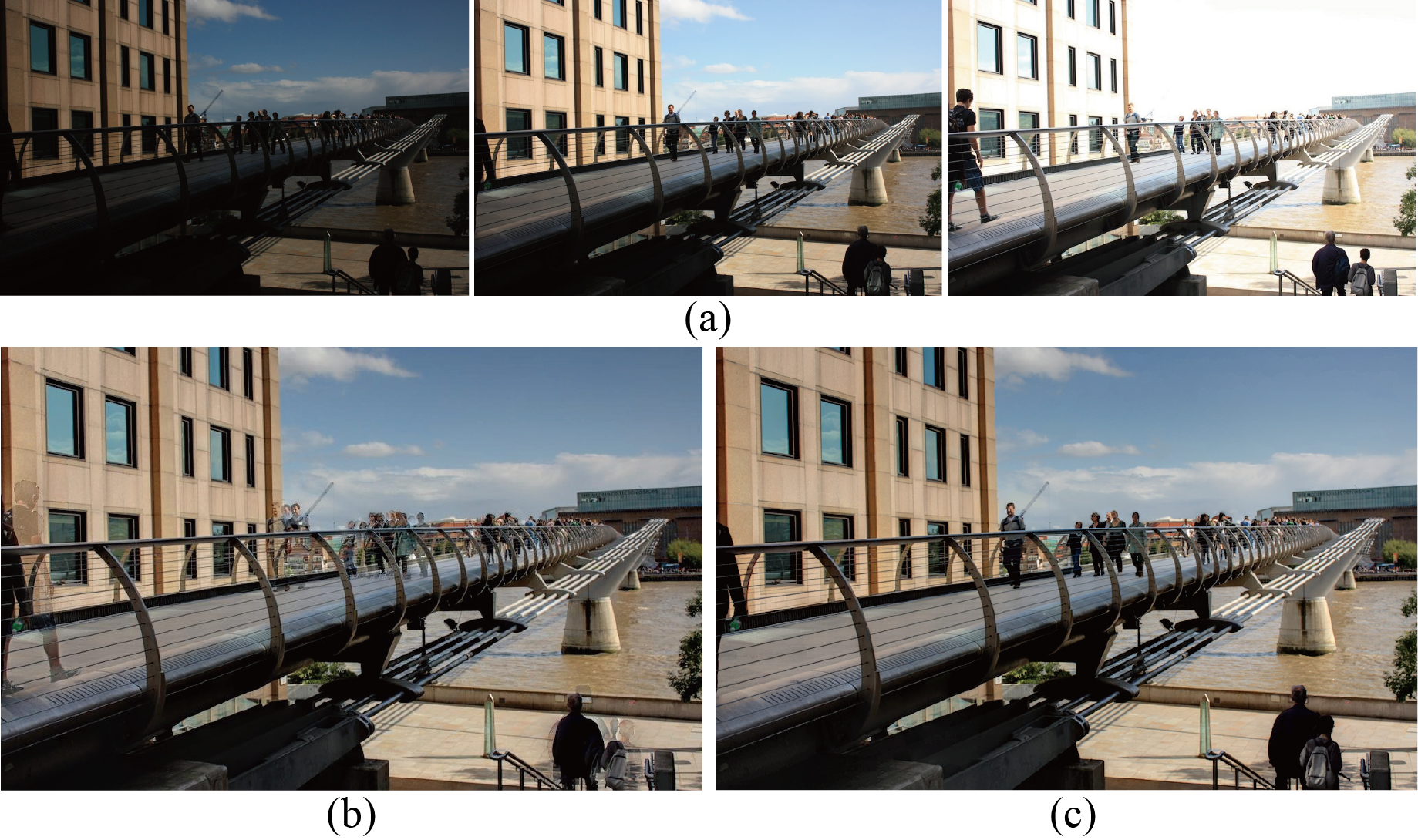}
	\caption{The ghosting artifacts of MEF method. (a) Input image sequence by courtesy of Pece~\cite{Pece2011Bitmap}. (b) Result of Mertens~\cite{Mertens2007Exposure}. (c) Our result.}
	\label{fig:intro}
\end{figure}

Since its first introduction in 1980's~\cite{Burt1984The}, image blending algorithms evolve quickly. They produce high-quality results when the input sequence is captured by static cameras mounted on a tripod under static scenes.
Mertens~\emph{et al.}~\cite{Mertens2007Exposure} combined contrast, saturation and exposedness information to generate weight maps and applied pyramid reconstruction to fuse multi-exposure image sequence. The method is robust and works well for static inputs. However, the weight maps are often too noisy and, sometimes,
may yield artifacts such as unnatural transitions and detail losses. Later, some modified exposure fusion methods~\cite{Li2012Fast,Li2013Image,Gu2012Gradient,Paul2016Multi} were
proposed to improve the performance of fusion image by filters~\cite{Li2012Fast,Li2013Image} or through gradient
reconstructions~\cite{Gu2012Gradient,Paul2016Multi}.

Above typical MEF algorithms require input exposures to be perfectly aligned. Otherwise, any motion, either due to dynamic
scenes or hand-shakes, will cause blurring/ghosting artifacts.
In particular, the method~\cite{Mertens2007Exposure} needs the inputs with strict alignment because every candidate pixel in the stack contributes to the final pixel value. If there are any misaligned regions, the fusion results would suffer from artifacts. As shown in Fig.~\ref{fig:intro}, the fused result of~\cite{Mertens2007Exposure} suffers from severe ghosting. The performance of fusion is highly dependent on the accuracy of motion estimation. Therefore, exploring efficient motion compensation strategies is essential.
According to Tursun~\emph{et al.}~\cite{Tursun2015The}, existing motion compensation methods can be divided into several categories: the moving objects removal~\cite{Zhang2010Gradient}, the moving objects selection~\cite{Wang2013An,Lee2014Ghost}, the optical flow based registration~\cite{Bogoni2000Extending,Zimmer2011Freehand,Kalantari2017Deep} and the patch-based registration~\cite{Sen2012Robust,Hu2012Exposure,Hu2013HDR,Ma2017Robust}.
Their core idea is to detect moving objects, then the dynamic areas are excluded or
assigned with small weights when synthesizing inputs.

In this paper, we propose a hybrid synthesis fusion method for hand-held camera inputs with dynamic contents. We first apply optical flow~\cite{Kroeger2016Fast} to align inputs to the reference. Optical flow methods are able to align images with complex motions, but produce deformations in the regions with no correspondences, which are caused by occlusions that either due to the parallax or owing to the dynamic contents. These artifacts usually affect the final fusion result during the merging process. As a result, we segment images into superpixels~\cite{Achanta2012SLIC} and use PatchMatch~\cite{barnes2009patchmatch} to correct error superpixels which are identified by computing flow motion variances. After the registration of optical flow and the compensation of PatchMatch, a fully registered image stack is delivered. Finally, we fuse registered images by~\cite{Mertens2007Exposure}. This hybrid synthesis brings many benefits to the fusion process. First, the fusion results are free from artifacts for a fully aligned input sequence. Second, we make full use of information from each input so as to generate a high-quality fused image. Third, the method not only possesses a high efficiency but also runs robustly on many challenging sequences. The main contribution of our method is that we combine the efficiency of flow-based algorithms and the accuracy of patch-based algorithms, therefore, generating higher quality fusion results than any flow-based methods or patch-based methods. We conduct comprehensive experiments to evaluate our method both qualitatively and quantitatively. The experimental results demonstrate the effectiveness and robustness of our approach.


\section{Related Works}

\subsection{MEF Algorithms with Static Inputs}
Burt~\emph{et al.} first applied Laplacian pyramid decomposition for image fusion~\cite{Burt1983The}. Then, some methods adopted the concept of pyramid decomposition to fuse images. They differ mainly in the definition of weight maps~\cite{Burt1993Enhanced,Mertens2007Exposure,shen2014exposure}.
Burt~\emph{et al.} defined a match measure and a salience measure as functions of the neighborhood of each pyramid~\cite{Burt1993Enhanced}.
Mertens~\emph{et al.} considered contrast, saturation and exposedness simultaneously to compute the weight maps~\cite{Mertens2007Exposure}. Shen~\emph{et al.} proposed a hybrid exposure weight that incorporates the local weight, global weight and saliency weight~\cite{shen2014exposure}. Some algorithms computed fusion parameters in the gradient domain and obtained the results by applying transformation techniques such as solving Poisson equation~\cite{Gu2012Gradient} or Haar wavelet~\cite{Paul2016Multi}. Edge preserving filters had been applied to retrieve fusion image edges, such as guided filter~\cite{Li2013Image,He2010Guided,Li2014Weighted}, bilateral filter~\cite{Tomasi1998Bilateral} and recursive filter~\cite{Li2012Fast}. These filter-based methods always possess faster speed.

The above-mentioned MEF algorithms are just suitable for static scenes. Some misalignments caused by dynamic scenes or hand-held cameras would lead to blurring/ghosting artifacts. For example, small movements of tree leaves in the wind would produce blurring results. To reduce such artifacts during fusion, a variety of deghosting algorithms have been proposed.

\subsection{Deghosting Algorithms with Dynamic Scenes}

\subsubsection{Optical Flow Registration}

Optical flow methods~\cite{Kroeger2016Fast,BOUGUET1999Pyramidal,Farneb2003Two,Baker2007A,Weinzaepfel2014DeepFlow} realize registration with pixel-level accuracy and are effective for aligning moving objects between two images. Bogoni~\emph{et al.} first applied optical flow to align multi-exposure images before fusion~\cite{Bogoni2000Extending}. They introduced a two-phase alignment strategy. The method first performs a global affine registration and then estimates the optical flow between input and reference. Zhang~\emph{et al.} proposed an edge-based image registration approach to guide the intensity-based registration which adopts optical flow estimation~\cite{Zhang2001A}. Kang~\emph{et al.} used gradient-based optical flow to compute a dense motion field and proposed a specialized HDR merging process to reject the artifacts of registration~\cite{Kang2003High}. Zimmer~\emph{et al.} presented an energy-based method for estimating the optical flow, which is robust in the presence of noise and occlusion~\cite{Zimmer2011Freehand}. Kalantari~\emph{et al.} applied optical flow to roughly align inputs and then reduced the artifacts of alignment during merging~\cite{Kalantari2017Deep}. The deep learning method~\cite{Kalantari2017Deep} owns sufficient prior information and generates results with good performance.

\subsubsection{Patch-based Registration}

Patch-based approaches aim to reconstruct ghost regions in output images by transferring information from inputs that are determined by patch matching.
The PatchMatch methods find matching patches from source to target and reconstruct source image~\cite{barnes2009patchmatch,barnes2010generalized,barnes2011patchmatch}.
Menzel and Guthe introduced a hierarchical blocking matching method that addresses both the camera and the scene motion~\cite{Menzel2007Freehand}. Sen~\emph{et al.} converted each image into a linear space by invoking the CRF~\cite{Sen2012Robust}. They proposed an energy minimization approach based PatchMatch. Hu~\emph{et al.} proposed a homography-based patching approach to handle the scene movement~\cite{Hu2012Exposure}. Then, Hu~\emph{et al.}'s recent work presented a PatchMatch based algorithm that deals with saturated regions and moving objects simultaneously~\cite{Hu2013HDR}. Ma~\emph{et al.} decomposed an image patch into three independent components and processed them respectively~\cite{Ma2017Robust}. Except for Ma's patch decomposition method~\cite{Ma2017Robust}, other patch-based methods are computationally costly due to the intensive searching and patching operations.

The intensity mapping function (IMF)~\cite{grossberg2003determining} has been used in either optical flow registration~\cite{Kalantari2017Deep} or patch-based registration~\cite{Hu2013HDR,Ma2017Robust} to map between intensity values of any two exposures. IMF is robust to scenes and camera motions, and can be implemented by modified histogram matching efficiently.

\subsection{Superpixel Segmentation}

\begin{figure}[t]
	\centering
	\includegraphics[width=1.0\linewidth]{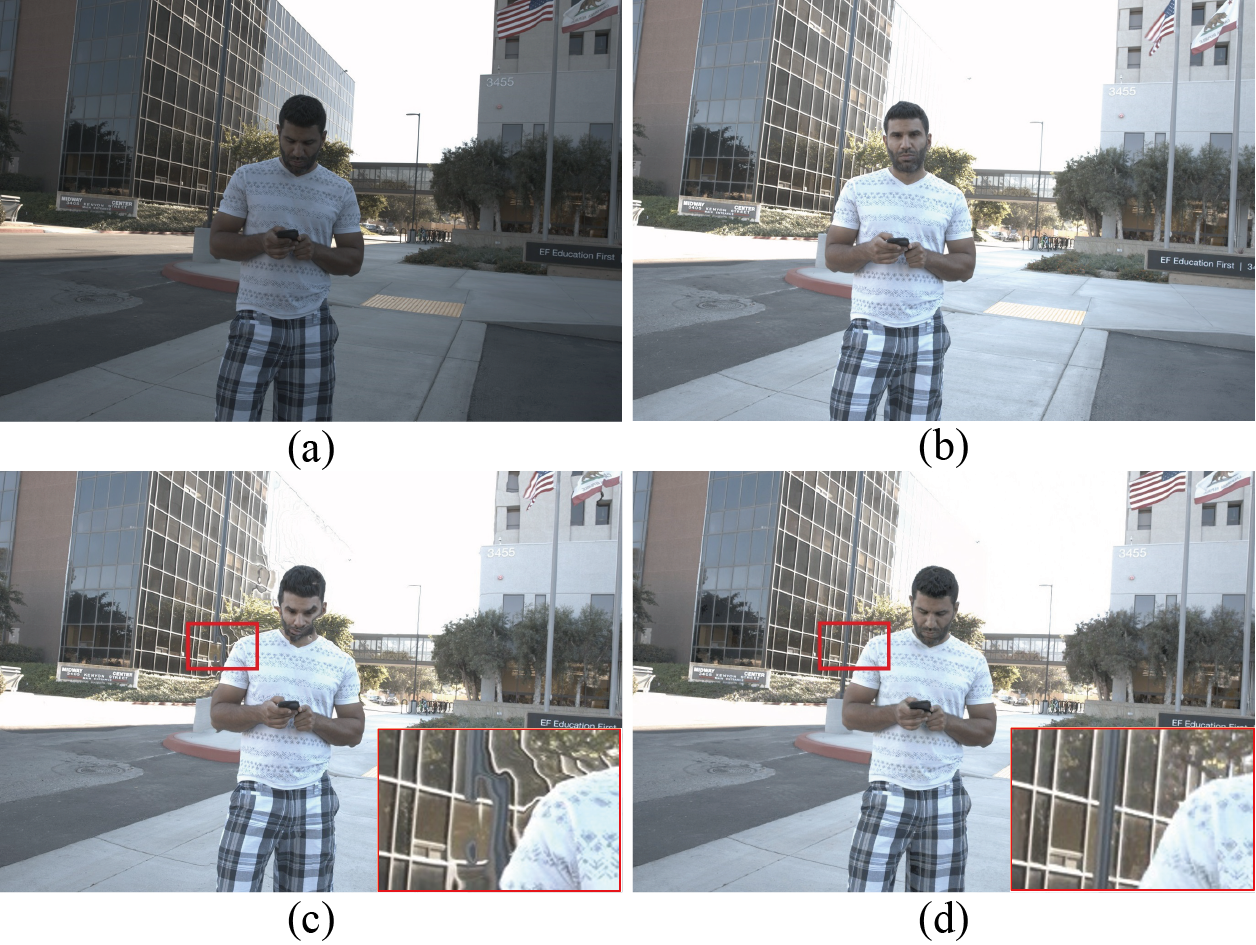}
	\caption{The image registration comparison between optical flow method and patch-based method when dealing with regions with no correspondences. (a) and (b) are input images by courtesy of Kalantari~\cite{Kalantari2017Deep} where (a) is the reference image. (c) Result of optical flow method~\cite{Kroeger2016Fast}. (d) Registration result of patch-based method~\cite{barnes2009patchmatch}.}
	\label{fig:depthchange}
\end{figure}

Superpixel segmentation methods divide an image into small regions which are rendered with uniform color, brightness and texture. Superpixels not only maintain the boundary information of objects but also provide benefits for subsequent image processing tasks. Felzenszwalb~\emph{et al.} proposed the most classic graph-based image segmentation algorithm~\cite{felzenszwalb2004efficient}. Achanta~\emph{et al.} proposed a simple linear iterative clustering (SLIC) superpixel algorithm which segments images to random regions with approximately same size~\cite{Achanta2012SLIC}, Bergh~\emph{et al.} introduced superpixels extracted via the energy-driven sampling (SEEDS) method based on a simple hill-climbing optimization. SEEDS divides image into superpixels with quite random sizes~\cite{Bergh2012SEEDS}, and Zheng~\emph{et al.} obtained approximately rectangular superpixels by linear spectral clustering (LSC)~\cite{Zhengqin2015Superpixel}. We select SLIC~\cite{Achanta2012SLIC} to segment the reference image.


\section{Motivation}
Optical flow has been extensively used for motion registration and motion analysis. With regards to the exposure fusion, several works utilize optical flow for image alignment during fusion results reconstruction~\cite{Bogoni2000Extending,Zimmer2011Freehand,Kalantari2017Deep}.
Optical flow methods detect pixel motions and then find an appropriate matching across different exposures. On the other hand, motions can also be compensated by patch matching strategies~\cite{Sen2012Robust,Hu2013HDR,kalantari2013patch}. Both of the flow-based and patch-based methods compensate motions not only from the cameras but also from dynamic objects. Ghosting effects can be easily introduced by poor pixel registrations. Therefore, the high-quality alignments ensure high-quality reconstructions.

It is important to disclose the strengths and weaknesses with respect to flow-based and patch-based methods. There are some methods apply PatchMatch for large displacement optical flow~\cite{bao2014fast,hu2016efficient}. Inspired by nearest neighbor field (NNF) algorithms, they blend random search strategies with the optical flow problems. As for exposure fusion, optical flow methods and patch-based methods have their own advantages and drawbacks to certain kinds of scenes~\cite{Tursun2015The,Kalantari2017Deep}. Before the presentation of our algorithm, we would like to analyze the two registration strategies through the following aspects.

\subsection{Discontinuous Depth}

If a scene is purely flat, or the scene objects are far away from the camera, the alignment can be easily realized by a single homography model. However, when depth changes exist in the scene, the homography is no longer valid~\cite{Hartley2003}. If depth variations were continuous, the mesh-based registration can achieve good results~\cite{Liu2013,lin2017MPA}. Nevertheless, if the variations were discontinuous, neither the homography nor the mesh warps produce satisfactory results.

Optical flow approaches possess a higher degree of freedom than mesh-based methods, leading to a higher registration accuracy in regions with continuous depth variations. However, optical flow methods often produce erroneous results when encountered with discontinuous depth, due to the absence of pixels caused by occlusions along the object boundary. The artifacts are unavoidable in such regions. In contrast, patch-based algorithms can overcome the challenge of no correspondences by filling the missing regions with similar patches in the reference image. They utilize the patch redundancy and synthesize missing pixels by patch matching algorithm.

Generally, patch-based methods achieve better results than optical flow
methods in the regions with discontinuous depth. A comparison of image registration between these two kinds of algorithms is shown in Fig.~\ref{fig:depthchange}. Here, we select a recent optical flow method~\cite{Kroeger2016Fast} and the classic patch-based method~\cite{barnes2009patchmatch} to conduct the comparison. We also adopt the two methods for subsequent experiments. Fig.~\ref{fig:depthchange}(c) and Fig.~\ref{fig:depthchange}(d) clearly indicate the correctness of our observation.

\begin{figure}[t]
	\centering
	\includegraphics[width=1.0\linewidth]{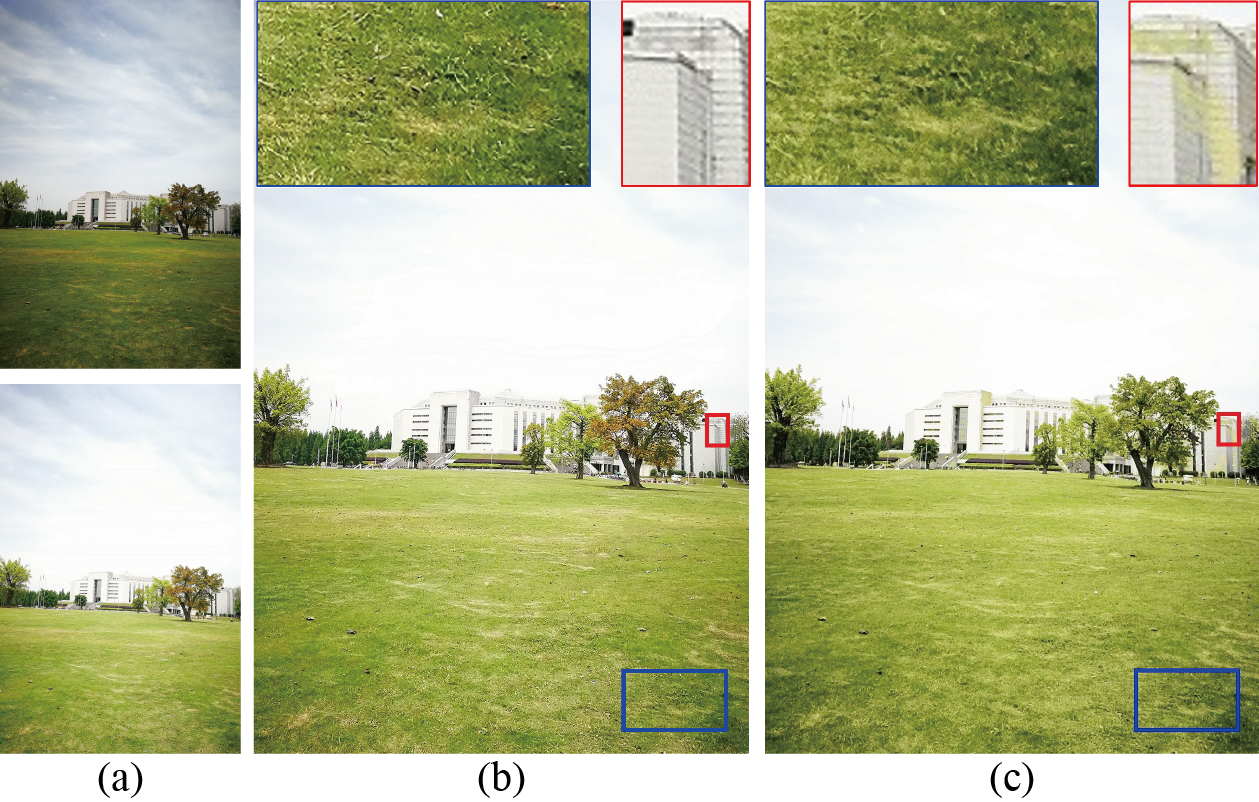}
	\caption{The comparison results of structured regions. (a) shows input images taken by us where the upper image is the reference image. (b) Result of optical flow method~\cite{Kroeger2016Fast}. (c) Registration result of patch-based method~\cite{barnes2009patchmatch}.}
	\label{fig:structural_regions}
\end{figure}

\begin{figure}[t]
	\centering
	\includegraphics[width=1.0\linewidth]{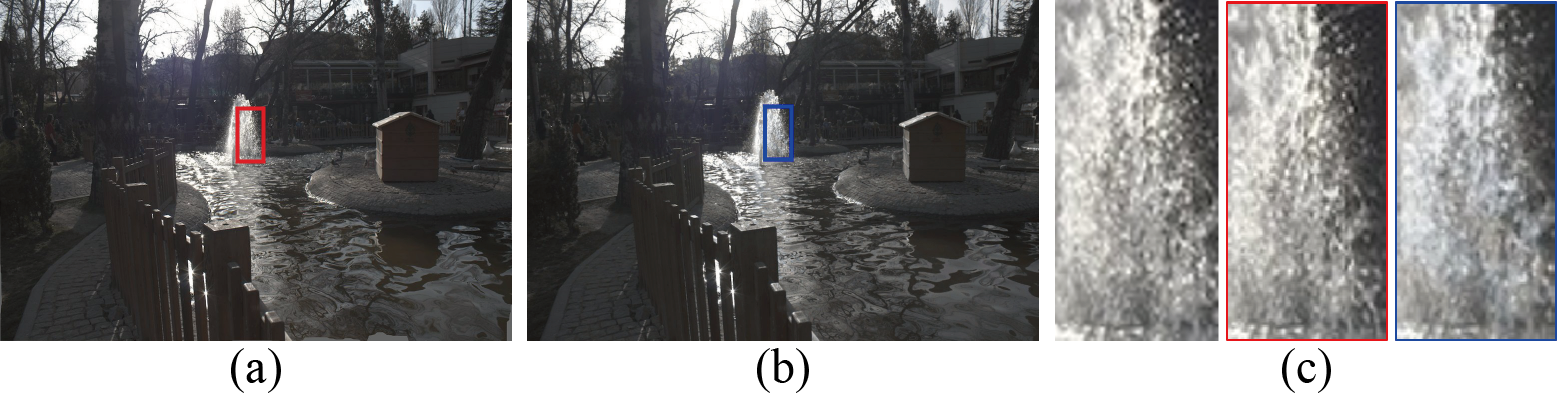}
	\caption{The comparison results of dynamic textures on Tursun~\emph{et al.}'s scene~\cite{tursun2016objective}. (a) Result of optical flow method~\cite{Kroeger2016Fast}. (b) Result of patch-based  method~\cite{barnes2009patchmatch}. (c) Details of reference image (left image in (c)), result by optical flow (red border) and result by patch match (blue border).}
	\label{fig:dynamictexture}
\end{figure}

\begin{figure*}[t]
	\centering
	\includegraphics[width=1.0\linewidth]{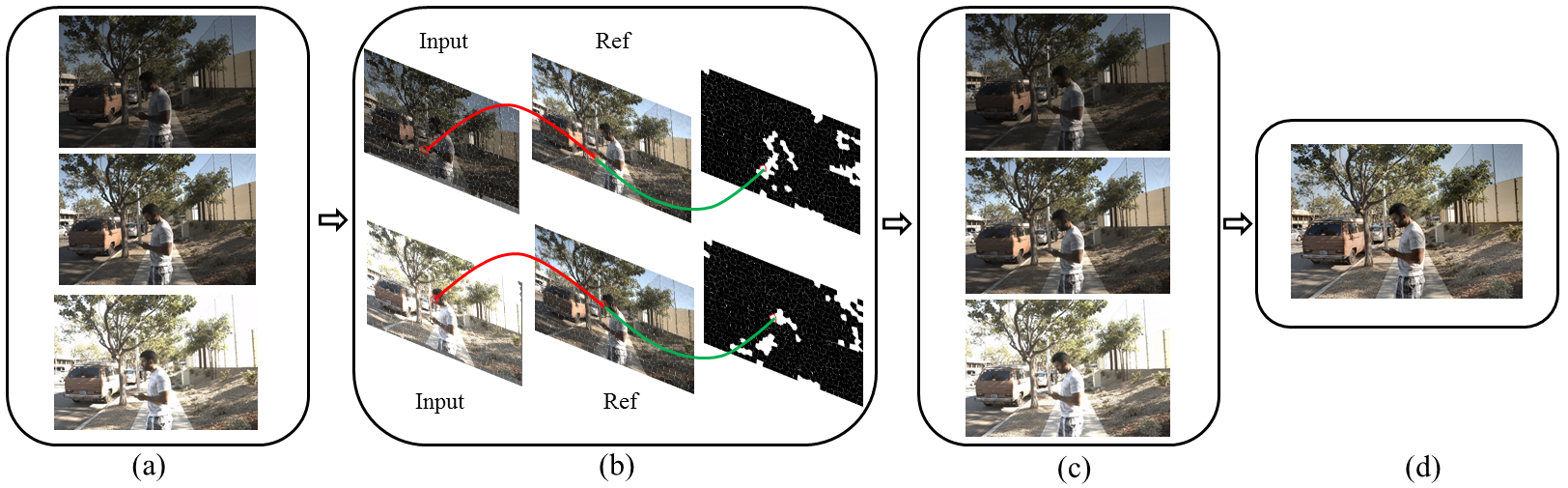}
	\caption{The pipeline of our method. (a) Input images with different exposures. (b) Optical flow results and error regions caused by error flows. (c) PatchMatch results. (d) Fusion results.}
	\label{fig:pipeline}
\end{figure*}

\subsection{Structured Regions}

It is challenging for patch-based methods to reconstruct inputs perfectly if the input image contains large structured regions with abundant details. Matching pixels is more preferable than matching patches in the structured regions. As shown in Fig.~\ref{fig:structural_regions}, rich texture information exists in the meadow regions and the building regions (the outline of ceramic tiles). Fig.~\ref{fig:structural_regions}(c) demonstrates that patch-based method may seek error regions from input to match the reference, leading to blur effects or error synthesis. On the contrary, optical flow performs well.

\subsection{Dynamic Textures}

Patch-based methods find patch correspondences and employ patch matching for the reconstruction. However, they can easily blur the areas that contain dynamic textures, e.g., fountains, sea-waves, flames and tree leaves in the wind. Because no accurate patches can be discovered due to the deformations of the textures.
Fig.~\ref{fig:dynamictexture} reveals the drawbacks of the patch-based method when handles the dynamic textures. The patch-based approach is not able to find suitable correspondences in the fountain regions (right image in Fig.~\ref{fig:dynamictexture}(c)). Different with the patch-based method, the result of optical flow method (middle image in Fig.~\ref{fig:dynamictexture}(c)) owns higher similarity to inputs due to its pixel-level optimization.

\subsection{Efficiency}

\begin{table}[t]
\small
\centering
\caption{Running time comparisons between optical flow methods and patch-based methods.}
\label{tab:runtime}
\resizebox{1.0\linewidth}{!}{
\begin{tabular}{c|c|cccc}
\toprule
                             & Image size         & \begin{tabular}[c]{@{}c@{}}Kroeger\\ \cite{Kroeger2016Fast}\end{tabular} & \begin{tabular}[c]{@{}c@{}}Farneb{\"a}ck\\ \cite{Farneb2003Two}\end{tabular} & \begin{tabular}[c]{@{}c@{}}Bouguet\\ \cite{BOUGUET1999Pyramidal}\end{tabular} & \begin{tabular}[c]{@{}c@{}}Barnes\\ \cite{barnes2009patchmatch}\end{tabular} \\ 
                             \midrule
\multirow{2}{*}{Time ($ms$)} & 900 $\times$ 600   & 50                                                                                        & 379                                                                                                            & 610                                                                                            & 2042                                                                                          \\
                             & 1500 $\times$ 1000 & 93                                                                                        & 1039                                                                                                           & 1173                                                                                           & 7694                                                                                          \\ 
\bottomrule
\end{tabular}
}
\end{table}

The speed of optical flow computation has been largely accelerated in the past several years. Patch matching methods can facilitate interactive image editing when reconstructing the whole image, however, they are still much slower than optical flow approaches even though applying the modern variation of PatchMatch~\cite{barnes2011patchmatch}.
We select several optical flow methods~\cite{Kroeger2016Fast,BOUGUET1999Pyramidal,Farneb2003Two} and the classic patch-based method~\cite{barnes2009patchmatch} to align images and record their running time accordingly. The results are summarized in Table~\ref{tab:runtime}.
Results indicate that a relatively new optical flow method~\cite{Kroeger2016Fast} has the fastest speed. Other optical flow methods~\cite{BOUGUET1999Pyramidal,Farneb2003Two} also match images in a short time, while patch-based method~\cite{barnes2009patchmatch} is much slower when compared with optical flow methods.

\subsection{Summary}

Table~\ref{tab:summary} summarizes the merits and limits with respect to the optical flow and patch-based methods. To design a good registration method, we want to combine their advantages.


\section{Our Method}

\begin{table}[!t]
\small
\centering
\caption{Pros and Cons of optical flow and patch match}
\label{tab:summary}
\begin{tabular}{lcc}
\toprule
  & Patch match  & Optical flow\\
\midrule
 A. Discontinuous depth  & $\surd$ &  \\
 B. Structured regions &  & $\surd$ \\
 C. Dynamic textures &  & $\surd$ \\
 D. Efficiency &  & $\surd$ \\
\bottomrule
\end{tabular}
\end{table}

Figure~\ref{fig:pipeline} shows our hybrid fusion pipeline. Taking efficiency into consideration, we employ optical flow to make a global registration and then some local problematic regions are corrected by PatchMatch. Without loss of generality, we take three images as an example. Fig.~\ref{fig:pipeline}(a) displays the input image sequence. Images are aligned to the reference image using optical flow~\cite{Kroeger2016Fast}. Fig.~\ref{fig:pipeline}(b) shows the process of detecting and handling error regions caused by optical flow. We segment the reference image using superpixels and calculate flow variance within each of them. If the variance exceeds a threshold, the corresponding superpixel is regarded as the error region. PatchMatch algorithm is applied to further align these error regions. Fig.~\ref{fig:pipeline}(c) shows results by PatchMatch correction. The combination of the two alignment methods not only achieves a fast speed, but also correctly aligns the challenging areas, such as regions containing occlusions and dynamic textures. Fig.~\ref{fig:pipeline}(d) is the final fusion results.

\subsection{Optical Flow Alignment}

Given a sequence of LDR images with low, medium, and high exposures, we first align the low and high exposure images to the medium exposure one (reference) using optical flow. We choose Kroeger~\emph{et al.}'s optical flow method~\cite{Kroeger2016Fast} as our coarse registration strategy, which is faster than other standard optical flow methods~\cite{BOUGUET1999Pyramidal,Farneb2003Two} for its efficient inverse search for patch correspondences. However, as illustrated in the previous section, optical flow methods will produce deformations in regions with discontinuous depth. If we directly fuse image stacks that are roughly aligned by optical flow, blur or ghosting artifacts are unavoidable. Locating these error regions and further aligning them is essential.

\begin{figure}[t]
	\centering
	\includegraphics[width=1.0\linewidth]{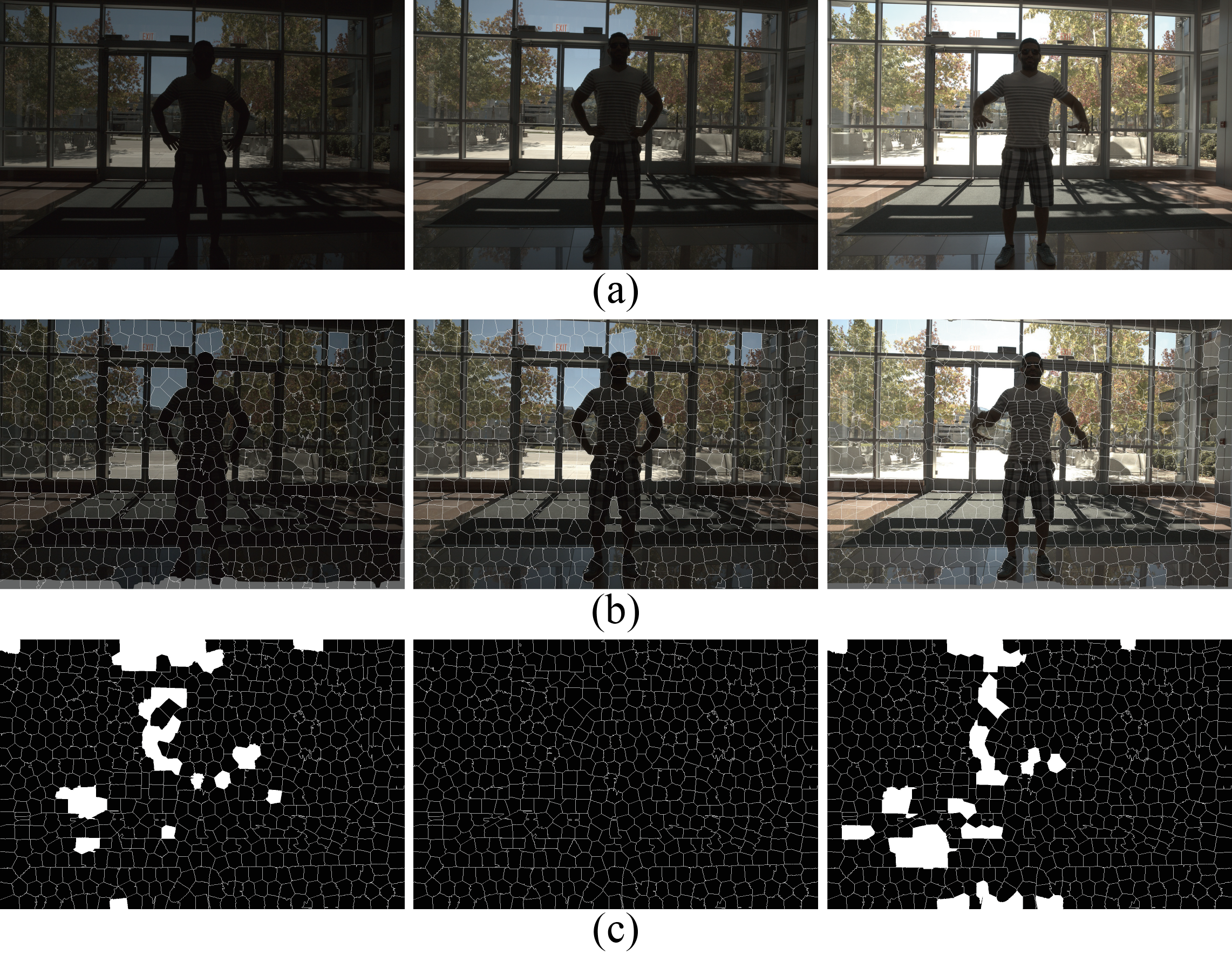}
	\caption{Results of error region detection. (a) Input images by courtesy of Kalantari~\cite{Kalantari2017Deep} with the middle one is regarded as the reference image. We apply superpixel segmentation on the reference image and reuse the segmentation mask for all images. (b) Warped results by optical flow. (c) Superpixels with incorrect alignment.}
	\label{fig:opticalflow}
\end{figure}

Instead of locating errors for every pixel, we segment images into superpixels, and detect errors for each superpixel. Dividing images into superpixels can not only maintain image continuity but also reduce the complexity of the subsequent image processing tasks. We apply SLIC segmentation~\cite{Achanta2012SLIC} to generate superpixels, which adapts a $k$-means clustering to efficiently generate superpixels. It owns the faster speed than LSC method~\cite{Zhengqin2015Superpixel} and has more regular superpixel shape than SEEDS method~\cite{Bergh2012SEEDS}. We first obtain the segmentation mask of the reference image by SLIC and then apply the mask to segment other aligned input images. As such, all input images share the same segmentation. When detecting error regions, we calculate flow motion variance for each superpixel. Error optical flow registration often possesses high motion variances. Therefore, applying a threshold can be adapted to distinguish correct superpixels and problematic superpixels. If the variance exceeds an empirical threshold ${T_{flow}}$, the corresponding region is labeled as error region:
\begin{equation}\label{eq:mask1}
\small
M = \left\{ {\begin{array}{*{20}{c}}
{0\hspace{.5cm}{V_{var}} < {T_{flow}}}\\
{1\hspace{.5cm}{V_{var}} > {T_{flow}}}
\end{array}} \right.
\end{equation}
where $M$ is error region mask with 1 represents incorrect region; ${{V_{var}}}$ is the motion variance of each superpixel. We calculate flow motion variances in horizontal (${V_{var  - x}}$) and vertical (${V_{var  - y}}$) directions, respectively. If one of them satisfies the condition, the mask is labeled as 1. Eq.~(\ref{eq:mask1}) is modified as:
\begin{equation}\label{eq:mask2}
\small
M = \left\{ {\begin{array}{*{20}{c}}
{0\hspace{.5cm}{V_{var  - x}} < {T_{flow}}\hspace{.12cm}or\hspace{.12cm}{V_{var  - y}} < {T_{flow}}}\\
{1\hspace{.5cm}{V_{var  - x}} > {T_{flow}}\hspace{.12cm}or\hspace{.12cm}{V_{var  - y}} > {T_{flow}}}
\end{array}} \right.
\end{equation}
Specifically, ${V_{var  - x}}$ is defined as:
\begin{equation}\label{eq:var}
\small
{V_{var  - x}} = \sqrt {\frac{{\sum\limits_{{v_x}} {{{({v_x} - av{g_x})}^2}} }}{N}}
\end{equation}
where ${{v_x}}$ represents horizontal motion values of each pixel; $N$ is total pixel number within each superpixel; ${av{g_x}}$ is the mean value of $N$ pixels. Replacing $x$ with $y$ in Eq.~(\ref{eq:var}), we can obtain ${V_{var  - y}}$.

Results of error region detection are shown in Fig.~\ref{fig:opticalflow}.
Fig.~\ref{fig:opticalflow}(a) displays the input image sequence where the second image is the reference. Fig.~\ref{fig:opticalflow}(b) shows alignment results by optical flow. Deformations appear in discontinuous regions such as the man's arm and the scene around his body. Fig.~\ref{fig:opticalflow}(c) exhibits the error region labels where white regions represent error regions. Please zoom in for a clearer observation and compare Fig.~\ref{fig:opticalflow}(b) and (d) for the validation.

\subsection{PatchMatch}

We adopt PatchMatch~\cite{barnes2009patchmatch} to correct error registrations. The algorithm offers substantial performance improvements for randomized patch-correspondence searching. We abandon running PatchMatch over the whole reference image by substitution of finding the nearest neighbor field (NNF) in a small region.
As shown in Fig.~\ref{fig:patchmatch}(c), we first get the height $h$ and width $w$ of one superpixel and then run PatchMatch in a region with height $2*h$ and width $2*w$ around the superpixel. Generally, running PatchMatch on the selected region can recover error registrations perfectly. However, the PatchMatch method tends to maintain color and structural similarity between input and reference. Directly applying PatchMatch to align error regions can eliminate the problem of dynamic objects but change the exposure information of inputs. It is unprofitable in exposure fusion.

\subsubsection{Illumination Adjustment}
We invoke IMF algorithm~\cite{grossberg2003determining} to adjust the exposure of the reference image to the input image before utilizing PatchMatch. IMF is capable of mapping between intensity values of any two exposures. The combination of IMF and PatchMatch reconstruction can exclude dynamic regions and maintain the exposure of inputs simultaneously.

\begin{figure}[t]
	\centering
	\includegraphics[width=1.0\linewidth]{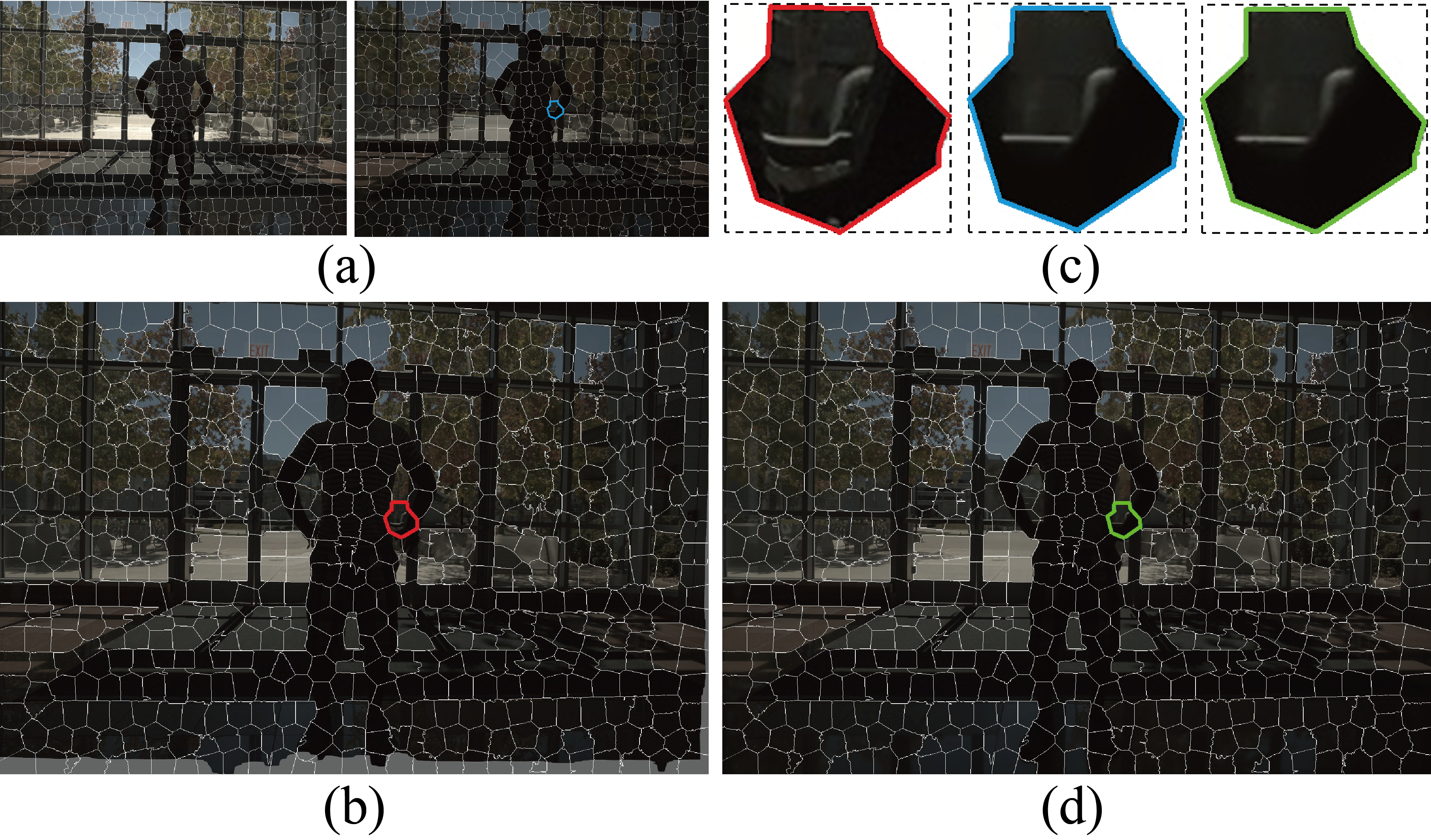}
	\caption{Results of PatchMatch on Kalantari~\emph{et al.}'s scene~\cite{Kalantari2017Deep}. (a) Reference image (left) and its exposure change result (right). (b) An optical flow result. (c) Some superpixel results. (d) PatchMatch result.}
	\label{fig:patchmatch}
\end{figure}

Fig.~\ref{fig:patchmatch}(a) shows the reference image $R$ (left) and its adjusted exposure image $L$ (right) by IMF. Fig.~\ref{fig:patchmatch}(b) is a warped result of optical flow. Fig.~\ref{fig:patchmatch}(c) highlights some superpixels: left superpixel (red border) originates from Fig.~\ref{fig:patchmatch}(b), which suffers distortions (the door frame shows obvious bend); middle superpixel (blue border) is the corresponding region from $L$, which is regarded as the reference image when employing PatchMatch; right superpixel (green border) is the PatchMatch result which is free from distortions. Fig.~\ref{fig:patchmatch}(d) is the final PatchMatch result which corrected all the problematic superpixels, leading to a high-quality registration to the reference image.

\subsubsection{Handling Noises}

\begin{figure*}[t]
	\centering
	\includegraphics[width=0.85\textwidth]{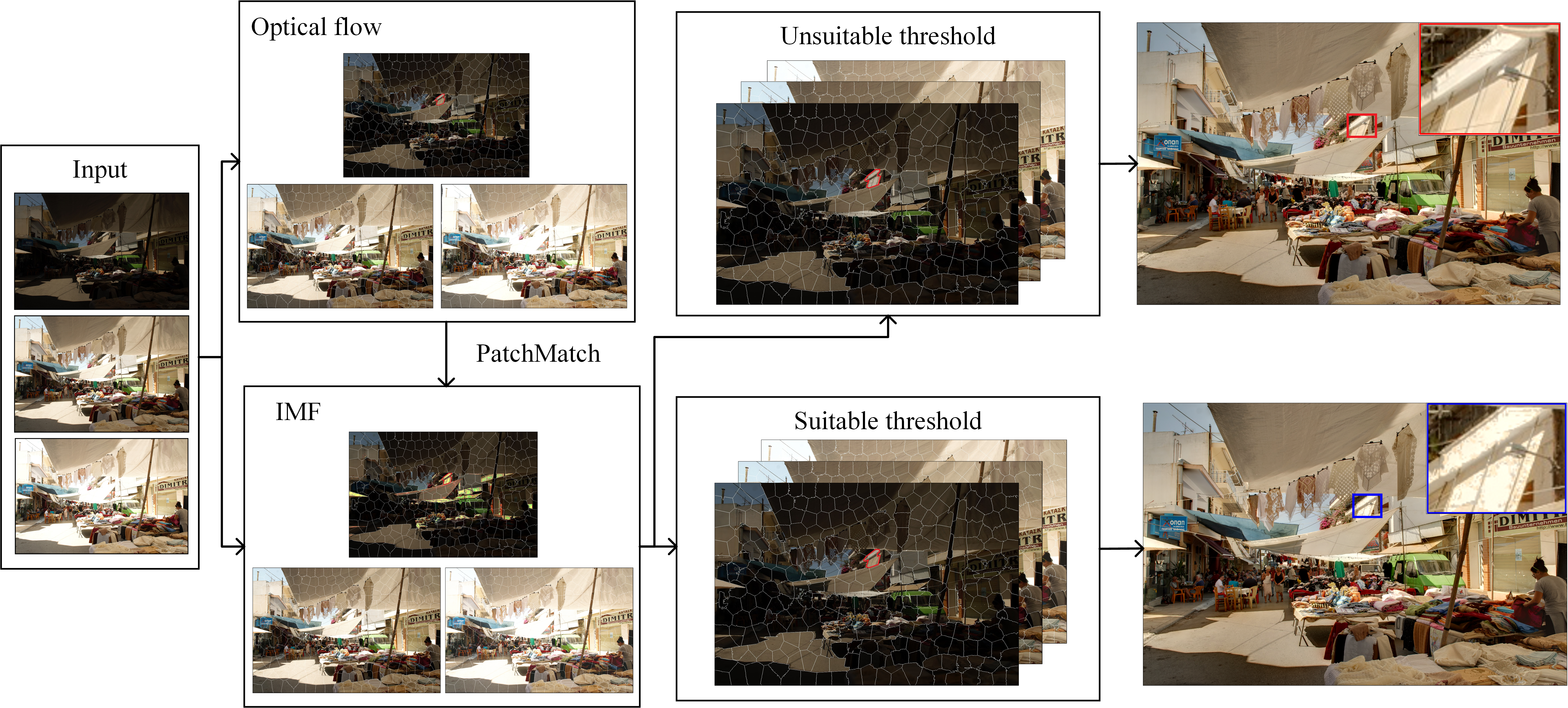}
	\caption{The influence of IMF. The problematic result (top row, right image) is influenced by IMF, while selecting the proper threshold can reduce such influence and generate good results (bottom row, right image).}
	\label{fig:IMF}
\end{figure*}

IMF algorithm shows how the radiometric response function can be related to the histograms of non-registered images with different exposures although the scene radiances are roughly constant. However, it is challenging for IMF to map image with high exposure to low exposure regions. Noises can be easily introduced in the mapped result. Fig.~\ref{fig:IMF} displays the influence of IMF. We select the middle image of inputs as the reference. In the first step, we obtain two groups of images: the optical flow results and the IMF results. Notice the noises in the left image of IMF results in Fig.~\ref{fig:IMF}. We further align some error superpixels from optical flow results to IMF results using PatchMatch. If an inappropriate threshold is selected, the noisy superpixel (red border in IMF result) is possibly considered as error superpixel. As a result, the noise will be brought to PatchMatch results for the PatchMatch's characteristic of maintaining structural similarity. The final fusion result will also be influenced.

In order to reduce such influence, we set different thresholds when detecting error regions. For under-exposed images, a higher threshold value ${T_{flow - u}}$ is preferred, which aims to align relatively few superpixels by PatchMatch. We found the noises generally appear in over-exposed regions without dynamic objects so it will not lead to producing ghosting in the final result. On the other hand, the selection of over-exposed threshold ${T_{flow - o}}$ is not as strict as under-exposed images. We display different conditions when $M$ is labeled as 1:
\begin{equation}\label{eq:diff-thres}
\small
\left\{ {\begin{array}{*{20}{c}}
{{V_{var }} > {T_{flow - u}} \hspace{.5cm} i < ref}\\
{{V_{var }} > {T_{flow - o}} \hspace{.5cm} i > ref}
\end{array}} \right.
\end{equation}

We arrange input images from low-exposure to high-exposure. $i<ref$ means image $i$ has lower exposure than reference image. Now a higher threshold ${T_{flow - u}}$ is selected. Whereas, when $i>ref$, ${T_{flow - o}}$ is selected.

\subsection{Implementation Details}

We summarize our approach in Algorithm~\ref{alg:our}. There are three parameters, number of superpixels ${N_s}$ and two threshold ${T_{flow - u}}$ and ${T_{flow - o}}$. The value of ${N_s}$ has a large impact on processing time. For an image with size 1500$\times$1000, we set ${N_s}$ around 580 empirically. Generally, it ranges from 560 to 600. Two threshold ${T_{flow - u}}$ and ${T_{flow - o}}$ control the number of error superpixels. We want to choose few regions from under-exposed images for the influence of noise caused by IMF, so we set a higher ${T_{flow - u}}$ value as 3.5. Whereas, we set ${T_{flow - o}}$ equals to 1.5 empirically.

\begin{algorithm}[t]
\small
\caption{Our algorithm}\label{alg:our}
\begin{algorithmic}[1]
\REQUIRE Source image sequence $\{ {S_k}\}  = \{ {S_k}|1 \le k \le K\} $
\STATE Select the reference ${S_r}$ and segment it into ${N_s}$ superpixels
\STATE Generate $K-1$ latent image $\{ {L_{k}}\}  = \{ {L_{k}}|1 \le k \le K,k \ne r\}$ using IMF
\FOR {each input image ${S_{k(k \ne r)}}$}
\STATE Align the input to the reference image using the optical flow
\STATE Calculate ${V_{var }}$ for each superpixel
\STATE Detect errors according to ${T_{flow}}$
\STATE Further align error regions to ${L_{k}}$ using PatchMatch
\ENDFOR
\STATE Fuse aligned image sequence $\{ \mathop {{S_k}}\limits^ \wedge  \}  = \{ \mathop {{S_k}}\limits^ \wedge  |1 \le k \le K\} $
\ENSURE Fusion result $\mathop S\limits^ \wedge$
\end{algorithmic}
\end{algorithm}


\section{Experiments}

In this section, we conduct various experiments to verify the performance of our method. We collect $24$ image sequences from previous publications, the Internet and our own captures, including daytime, nighttime, indoor and outdoor scenes, as well as static and dynamic scenes. Detailed information regarding the inputs is given in Table~\ref{tab:pic}.

Both objective evaluation and visual comparisons are conducted in the experiments.
In the objective assessment section, we compare image quality metric values and calculation complexities of our method against several state-of-the-art techniques which bear some resemblances with our method. Specifically, we compare against two patch-based methods of Sen~\emph{et al.}~\cite{Sen2012Robust} and Hu~\emph{et al.}~\cite{Hu2013HDR}, the patch decomposition method of Ma~\emph{et al.}~\cite{Ma2017Robust}, and the deep learning flow-based approach of Kalantari~\emph{et al.}~\cite{Kalantari2017Deep}. In the visual comparison section, we not only compare our approach with the above four methods of dynamic scenes but also select several representative MEF algorithms for static scene comparisons. The combating MEF algorithms are chosen to cover a diversity of types, including the classic exposure fusion method of Mertens~\emph{et al.}~\cite{Mertens2007Exposure}, the generalized random walks method of Shen~\emph{et al.}~\cite{Shen2011Generalized}, two filter-based method of Li~\emph{et al.}~\cite{Li2012Fast} and Li~\emph{et al.}~\cite{Li2013Image}, a Laplacian pyramid boosting method of Shen~\emph{et al.}~\cite{shen2014exposure}, a gradient-based method of Paul~\emph{et al.}~\cite{Paul2016Multi}, and the objective quality measure optimizing method of Ma~\emph{et al.}~\cite{ma2018multi}.

\begin{table*}[!ht]
\centering
\caption{Information about input image sequences}
\label{tab:pic}
\begin{minipage}[c]{0.45\textwidth}
\centering
\subtable[]{
\resizebox{0.8\linewidth}{!}{
\begin{tabular}{ccc}
\toprule
Source  & Size                       & Origin      \\
\midrule
Colosseum & 968$\times$648$\times$3   & \cite{Hu2013HDR}                 \\
Dome      & 968$\times$648$\times$3   & \cite{Hu2013HDR}                 \\
Duke      & 968$\times$648$\times$3   & \cite{Hu2013HDR}                 \\
Garden    & 968$\times$648$\times$3   & \cite{Hu2013HDR}                 \\
Happy     & 968$\times$648$\times$3   & \cite{Hu2013HDR}                 \\
Lady      & 968$\times$648$\times$3   & \cite{Hu2013HDR}                 \\
Barbeque  & 1500$\times$1000$\times$3 & \cite{Kalantari2017Deep} \\
Stand  & 1500$\times$1000$\times$3 & \cite{Kalantari2017Deep} \\
Train19   & 1500$\times$1000$\times$3 & \cite{Kalantari2017Deep} \\
Train51   & 1500$\times$1000$\times$3 & \cite{Kalantari2017Deep} \\
Capt3  & 1500$\times$1000$\times$3 & Our capture            \\
Capt4  & 1500$\times$1000$\times$3 & Our capture            \\
\bottomrule
\end{tabular}
}
}
\end{minipage}
\qquad
\begin{minipage}[c]{0.45\textwidth}
\centering
\subtable[]{
\resizebox{0.8\linewidth}{!}{
\begin{tabular}{ccc}
\toprule
Source   & Size                       & Origin      \\
\midrule
Pedestrian       & 1024$\times$682$\times$9   & \cite{tursun2016objective}\\
Fountain          & 1024$\times$682$\times$9   & \cite{tursun2016objective} \\
Noise             & 640$\times$480$\times$10   & \cite{gallo2009artifact}\\
Cafe              & 1662$\times$1088$\times$7  & EmpaMT\protect\footnotemark[2]\\
Forth2            & 1662$\times$1088$\times$7  & EmpaMT\\
Falls             & 1662$\times$1088$\times$7  & EmpaMT\\
Sunset            & 1662$\times$1088$\times$7  & EmpaMT\\
Zurich             & 1662$\times$1088$\times$7  & EmpaMT\\
Zurich2            & 1662$\times$1088$\times$7  & EmpaMT\\
Train1            & 1500$\times$1000$\times$5  & \cite{Kalantari2017Deep}\\
Capt1          & 1500$\times$1000$\times$5  & Our capture\\
Capt2          & 1500$\times$1000$\times$5  & Our capture\\
\bottomrule
\end{tabular}
}
}
\end{minipage}
\end{table*}

In order to make fair comparisons, we collect the codes from the authors of the above algorithms to generate their results with default settings.
\cite{Sen2012Robust}, ~\cite{Hu2013HDR} and~\cite{Kalantari2017Deep} methods can not generate comparison results directly. According to~\cite{Hu2013HDR}, we fuse their output aligned image stacks by~\cite{Mertens2007Exposure}. As illustrated in~\cite{Sen2012Robust} and~\cite{Kalantari2017Deep}, we generate their HDR outputs and obtain final comparison results using Photomatix software\footnote[1]{https://www.hdrsoft.com/index.html}.
\footnotetext[2]{http://empamedia.ethz.ch/hdrdatabase/index.php}

\subsection{Objective Assessment}

\subsubsection{Objective Evaluation Values}

\begin{table*}[t]
\small
\centering
\caption{Performance comparisons of our method with Sen~\emph{et al.}~\cite{Sen2012Robust}, Hu~\emph{et al.}~\cite{Hu2013HDR} and Ma~\emph{et al.}~\cite{Ma2017Robust}. The quality value ranges from 0 to 1 with a higher value indicating better quality.}
\label{tab:objective_value1}
\resizebox{0.95\textwidth}{!}{
\begin{tabular}{c|cccccccccccc|c}
\toprule
          & Pedestrian     & Fountain   & Noise    & Cafe           & Forth2         & Falls     & Sunset         & Zurich          & Zurich2         & Train1         & Capt1            & Capt2            & Avg            \\
\midrule
\cite{Sen2012Robust}           & 0.731          & 0.573     & 0.265     & 0.684          & 0.794          & 0.589          & 0.688          & 0.745          & 0.729          & 0.619          & 0.486          & 0.476          & 0.615          \\
\cite{Hu2013HDR}          & 0.766           & 0.635   & 0.753       & 0.751          & 0.825          & 0.767          & 0.759          & 0.744          & 0.782          & 0.619          & \textbf{0.499} & 0.480          & 0.698          \\
\cite{Ma2017Robust}  & 0.777          & 0.616     & \textbf{0.783}     & 0.725          & 0.818          & 0.717          & 0.730          & \textbf{0.771} & 0.739          & \textbf{0.639} & 0.468          & 0.448          & 0.686          \\
Our            & \textbf{0.781}   & \textbf{0.637} &0.769 & \textbf{0.767} & \textbf{0.845} & \textbf{0.781} & \textbf{0.802} & 0.754          & \textbf{0.799} & 0.620          & 0.490          & \textbf{0.486} & \textbf{0.711}\\
\bottomrule
\end{tabular}
}
\end{table*}

\begin{table*}[t]
\small
\centering
\caption{Performance comparisons of our method with Kalantari~\emph{et al.}~\cite{Kalantari2017Deep}. The quality value ranges from 0 to 1 with a higher value indicating better quality.}
\label{tab:objective_value2}
\resizebox{0.95\textwidth}{!}{
\begin{tabular}{c|cccccccccccc|c}
\toprule
    & Colosseum & Dome  & Duke  & Garden & Happy & Lady  & Barbeque & Stand & Train19 & Train51 & Capt3   & Capt4   & Avg   \\
\midrule
\cite{Kalantari2017Deep}   & 0.343     & 0.559 & 0.477 & 0.181  & 0.441     & 0.443 & 0.303       & 0.554       & 0.329   & 0.469   & 0.478 & 0.388 & 0.414 \\
Our & \textbf{0.457}     & \textbf{0.656} & \textbf{0.581} & \textbf{0.461}  & \textbf{0.585}     & \textbf{0.549} & \textbf{0.487}       & \textbf{0.783}       & \textbf{0.491}   & \textbf{0.614}   & \textbf{0.608} & \textbf{0.602} & \textbf{0.573}\\
\bottomrule
\end{tabular}
}
\end{table*}

\begin{table}[!t]
\small
\centering
\caption{Average execution time in seconds on 12 source image sequences of size 1500$\times$1000$\times$5.}
\label{tab:complexities}
\resizebox{1.0\linewidth}{!}{
\begin{tabular}{c|ccccc}
\toprule
 Algorithm & Kalantari\cite{Kalantari2017Deep} &  Hu\cite{Hu2013HDR} & Sen\cite{Sen2012Robust}  &  Ma\cite{Ma2017Robust}  & Our  \\
\midrule
Time(s)    & 68$ \pm $5.1     &  381$ \pm $38    & 252$ \pm $20 & 16$ \pm $0.7 & 12$ \pm $1\\
\bottomrule
\end{tabular}
}
\end{table}

There are many objective assessments for multi-exposure image fusion algorithms~\cite{Liu2012Objective}.
We adopt a popular evaluation metric ${Q_S}$~\cite{Piella2003A} as our evaluation criteria. Piella and Heijmans defined the quality metric ${Q_S}$ based on Wang's UIQI method~\cite{Wang2002A}. Wang's method defined image quality index ${Q_0}$ as:
\begin{equation}\label{eq:Q_0}
\small
{Q_0}(a,b) = \frac{1}{{|W|}}\sum\limits_{w \in W} {{Q_0}(a,b|w)}
\end{equation}
where $a$ and $b$ are input images; ${{Q_0}(a,b|w)}$ is computed for the values $a(i,j)$ and $b(i,j)$ where pixels $(i,j)$ lie in the sliding window $w$; $W$ is the family of all windows and $|W|$ is the cardinality of $W$.

Piella~\emph{et al.} modified ${{Q_0}(a,b)}$ to ${{Q_S}(a,b,f)}$ to express the quality of the fused image $f$ given by the inputs:
\begin{equation}\label{eq:Qs}
\small
{Q_S} = \frac{1}{{|W|}}\sum\limits_{w \in W} {[\lambda (w){Q_0}(a,f|w) + (1 - \lambda (w)){Q_0}(b,f|w)]}
\end{equation}
where ${\lambda (w)}$ is a local weight between 0 and 1 indicating the relative importance of images $a$ compared to image $b$. It is defined as:
\begin{equation}\label{eq:lamda}
\small
\lambda (w) = \frac{{s(a|w)}}{{s(a|w) + s(b|w)}}
\end{equation}
where ${s(a|w)}$ and ${s(b|w)}$ are the local saliency which reflects the local relevance of input within the window $w$.

The quality metric evaluates how much of the salient information contained in each of the input images has been transferred into the fused image. It ranges from 0 to 1 with higher values indicating better performance. The comparison results on 24 sequences are shown in Table~\ref{tab:objective_value1} and Table~\ref{tab:objective_value2}. Note that we divide the results into two tables because the deep learning method~\cite{Kalantari2017Deep} is just suitable for the condition with three inputs. Table~\ref{tab:objective_value1} displays the comparisons of our method with Sen~\emph{et al.}~\cite{Sen2012Robust}, Hu~\emph{et al.}~\cite{Hu2013HDR} and Ma~\emph{et al.}~\cite{Ma2017Robust}. As can be seen, our method provides results with better ${Q_S}$ values in most of the cases. Table~\ref{tab:objective_value2} shows the comparison results with Kalantari~\emph{et al.}~\cite{Kalantari2017Deep}. Although the deep learning method owns more prior information, our method preserves more information from the source image to the fusion image.

\subsubsection{Calculation Complexity}

Besides the fusion performance, computing efficiency is also important. We conduct a complexity comparison in Table~\ref{tab:complexities}.
Upon the same inputs, all experiments are conducted on a computer with i7 3.4 GHz CPU and 32G RAM. Our method takes approximately 12 seconds. Specifically, when coping with five inputs with size 1500$\times$1000, the optical flow needs 0.4 seconds; the superpixel segmentation takes 0.9 seconds; the fusion process needs 0.5 seconds; and the rest of time is spent on detecting error regions and using PatchMatch for synthesis. The PatchMatch method is time-consuming, so we only select a small number of erroneous areas to apply patch-based registration. From Table~\ref{tab:complexities}, we know that the proposed method is more computationally efficient. It is a little bit faster than Ma~\cite{Ma2017Robust} and much faster than Sen~\cite{Sen2012Robust}, Hu~\cite{Hu2013HDR} and Kalantari~\cite{Kalantari2017Deep}.

\subsection{Visual Comparisons}

\begin{figure*}[t]
	\centering
	\includegraphics[width=1.0\linewidth]{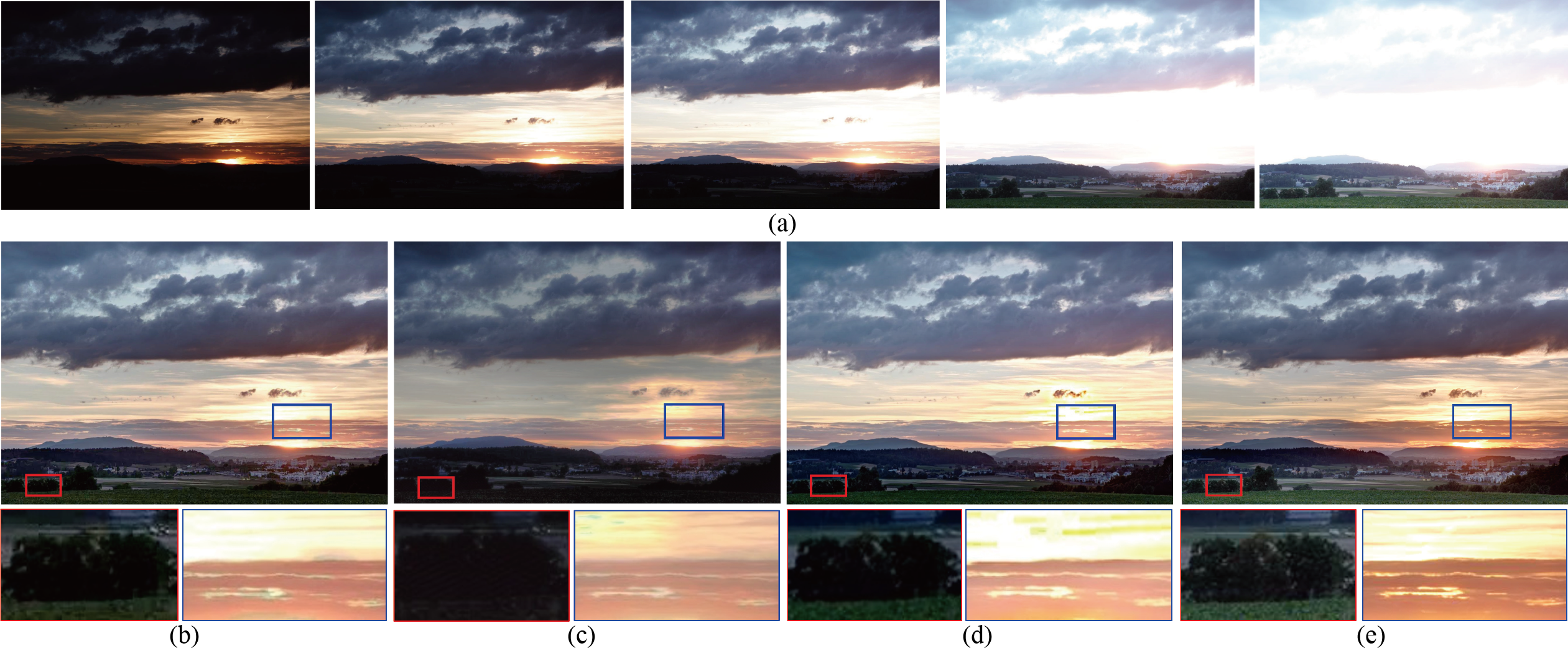}
	\caption{Comparisons of our method with Hu~\emph{et al.}~\cite{Hu2013HDR}, Sen~\emph{et al.}~\cite{Sen2012Robust}, Ma~\emph{et al.}~\cite{Ma2017Robust} with 5 inputs. (a) Source image sequence by courtesy of EmpaMT dataset. (b) Result of~\cite{Hu2013HDR}. (c) Result of~\cite{Sen2012Robust}. (d) Result of~\cite{Ma2017Robust} (e) Our result. Please refer to the supplement file for more comparisons and our results.}
	\label{fig:comp1}
\end{figure*}

\begin{figure*}[!t]
	\centering
	\includegraphics[width=1.0\linewidth]{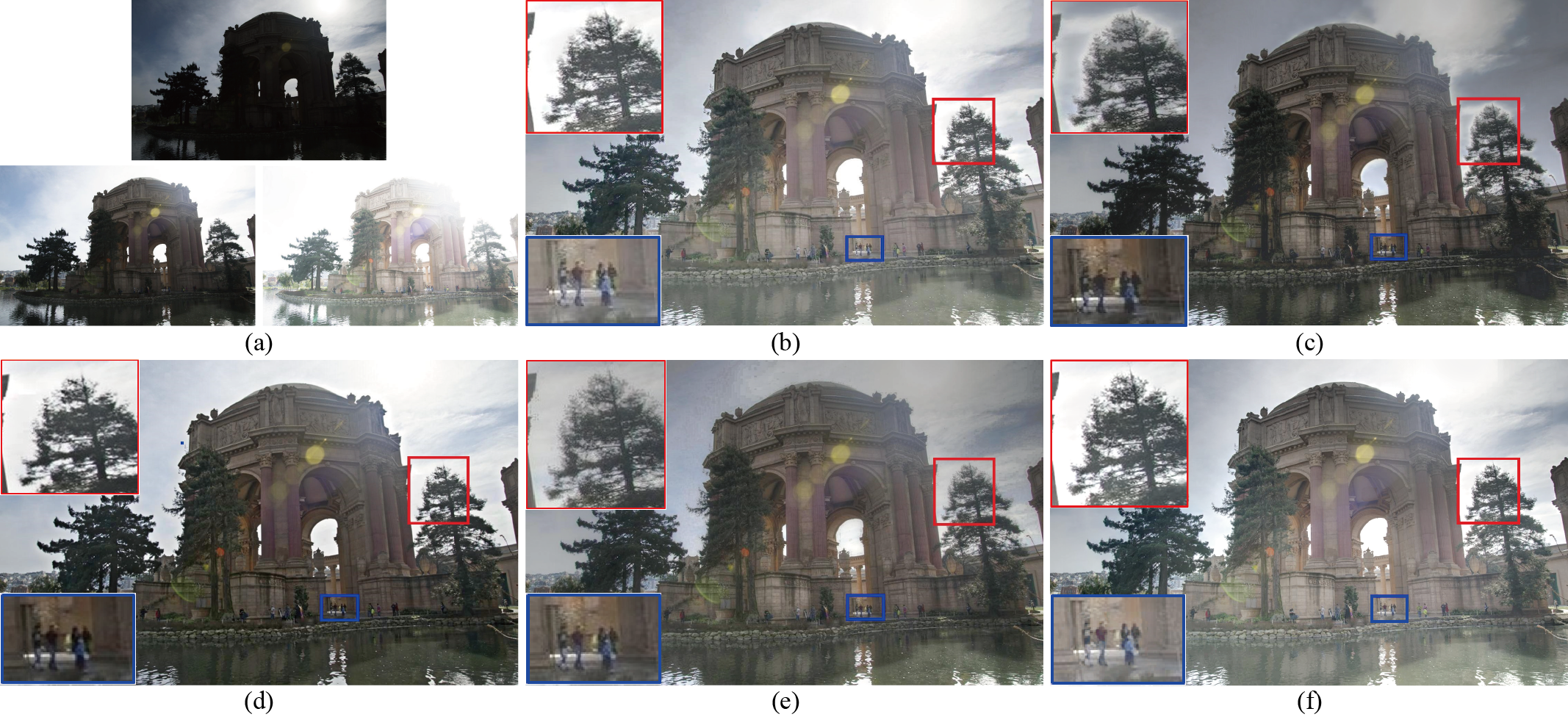}
	\caption{Comparisons of our method with Hu~\emph{et al.}~\cite{Hu2013HDR}, Sen~\emph{et al.}~\cite{Sen2012Robust}, Ma~\emph{et al.}~\cite{Ma2017Robust} and Kalantari~\emph{et al.}~\cite{Kalantari2017Deep} with 3 inputs. (a) Source image sequence by courtesy of Hu~\cite{Hu2013HDR}. (b) Result of~\cite{Hu2013HDR}. (c) Result of~\cite{Sen2012Robust}. (d) Result of~\cite{Ma2017Robust} (e) Result of~\cite{Kalantari2017Deep}. (f) Our result. All these five methods handle moving objects well and avoid ghosting effectively.}
	\label{fig:comp2}
\end{figure*}

\begin{figure*}[t]
	\centering
	\includegraphics[width=1.0\linewidth]{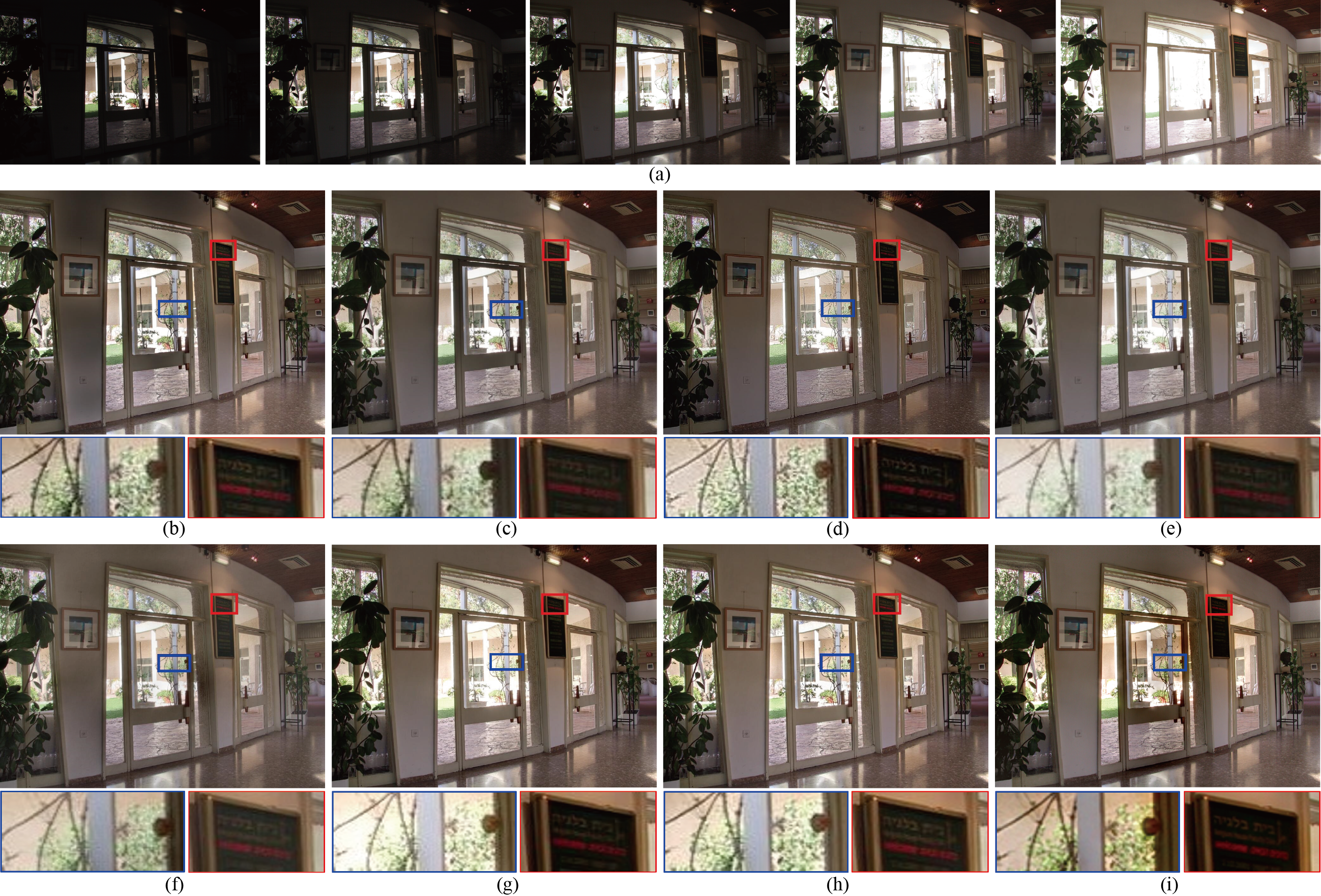}
	\caption{Comparisons with Li~\emph{et al.}~\cite{Li2012Fast}, Li~\emph{et al.}~\cite{Li2013Image}, Shen~\emph{et al.}~\cite{shen2014exposure}, Shen~\emph{et al.}~\cite{Shen2011Generalized}, Paul~\emph{et al.}~\cite{Paul2016Multi}, Mertens~\emph{et al.}~\cite{Mertens2007Exposure} and Ma~\emph{et al.}~\cite{ma2018multi}. (a) Source images by courtesy of Fattal~\cite{Fattal2002Gradient}. (b) Result of~\cite{Li2012Fast}. (c) Result of~\cite{Li2013Image}. (d) Result of~\cite{shen2014exposure}. (e) Result of~\cite{Shen2011Generalized}. (f) Result of~\cite{Paul2016Multi}. (g) Result of~\cite{Mertens2007Exposure}. (h) Result of~\cite{ma2018multi}. (i) Our result. }
	\label{fig:comp3}
\end{figure*}

\begin{figure}[t]
	\centering
	\includegraphics[width=1.0\linewidth]{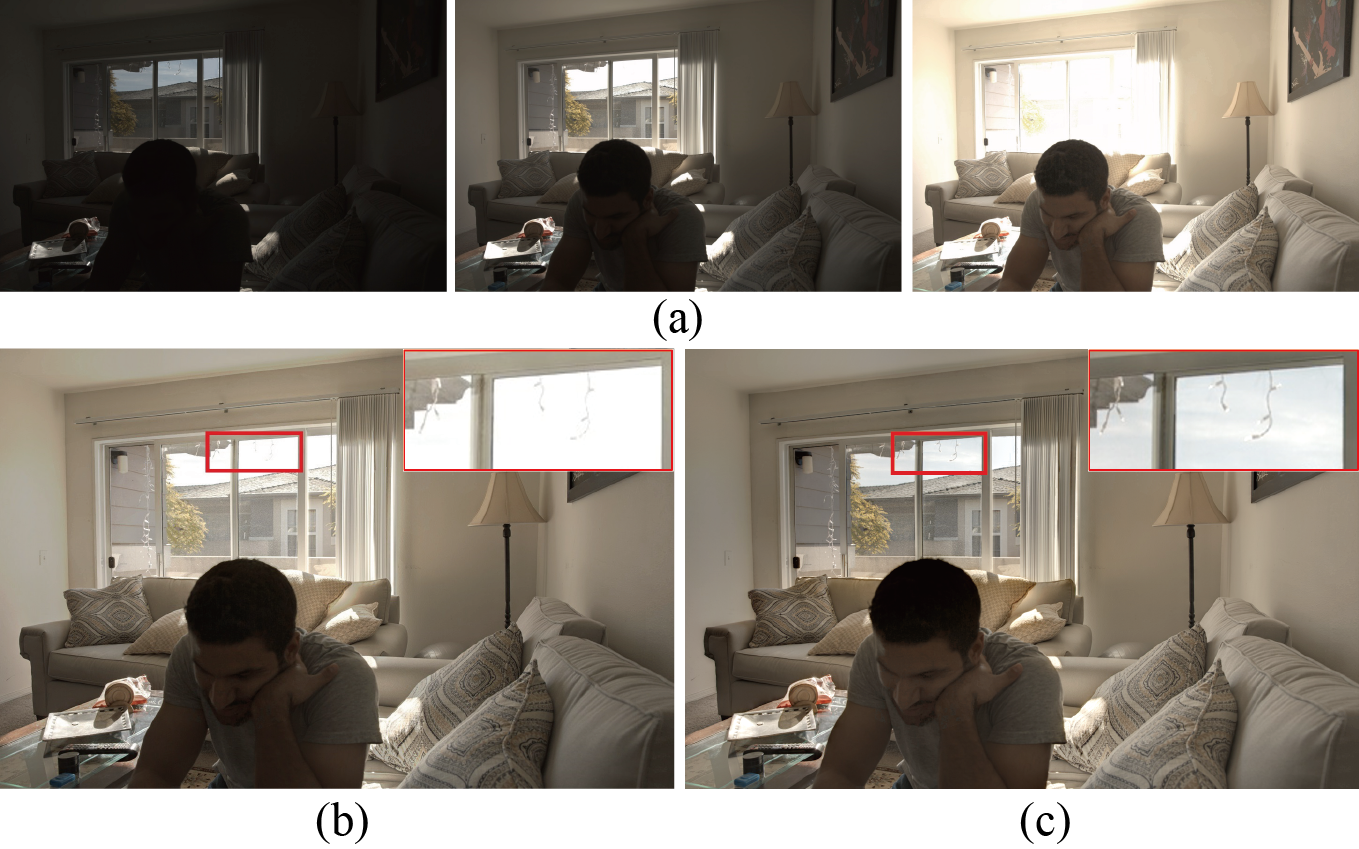}
	\caption{The comparison results of exposure fusion on Kalantari~\emph{et al.}'s scene ~\cite{Kalantari2017Deep}. (a) Aligned image sequence. (b) Result of Mertens~\cite{Mertens2007Exposure} implemented by OpenCV. (c) Our fusion result.}
	\label{fig:exposurefusion}
\end{figure}

In this part, we first compare our results with the above four methods: Sen~\emph{et al.}~\cite{Sen2012Robust}, Hu~\emph{et al.}~\cite{Hu2013HDR},  Ma~\emph{et al.}~\cite{Ma2017Robust} and Kalantari~\emph{et al.}~\cite{Kalantari2017Deep}. We show the comparison results in Fig.~\ref{fig:comp1} and Fig.~\ref{fig:comp2}.
Fig.~\ref{fig:comp1} shows the comparison results of our method with~\cite{Hu2013HDR},~\cite{Sen2012Robust} and~\cite{Ma2017Robust} with 5 inputs.
Fig.~\ref{fig:comp2} shows the comparison results of our method with~\cite{Hu2013HDR},~\cite{Sen2012Robust},~\cite{Ma2017Robust} and~\cite{Kalantari2017Deep} with 3 inputs. From Fig.~\ref{fig:comp1}(b) and Fig.~\ref{fig:comp2}(b), we found that the patch-based method~\cite{Hu2013HDR} can not deal with the structural regions well. Similar to~\cite{Hu2013HDR}, Sen's method~\cite{Sen2012Robust} also generate unsatisfactory results (dark regions in Fig.~\ref{fig:comp1}(c) and slight blurry trunks in Fig.~\ref{fig:comp2}(c)). The errors in motion estimation are difficult to avoid in the presence of tiny random motion for patch-based methods although they generally produce relatively good results. Ma's patch decomposition method~\cite{Ma2017Robust} can not recover some information perfectly such as the left tree in Fig.~\ref{fig:comp2}(d) because the regions in the reference image are under-exposed and the structures cannot be properly inferred from other exposures. Then, from Fig.~\ref{fig:comp2}(e), we found the flow-based deep learning method~\cite{Kalantari2017Deep} suffers from some artifacts in tree leaves and sky regions. Our method is free from such artifacts.

Then we select several typical MEF algorithms to compare: Li~\emph{et al.}~\cite{Li2012Fast}, Li~\emph{et al.}~\cite{Li2013Image}, Shen~\emph{et al.}~\cite{shen2014exposure}, Shen~\emph{et al.}~\cite{Shen2011Generalized}, Paul~\emph{et al.}~\cite{Paul2016Multi}, Mertens~\emph{et al.}~\cite{Mertens2007Exposure} and Ma~\emph{et al.}~\cite{ma2018multi}. They require static inputs without shaky scenes and dynamic objects. Fig.~\ref{fig:comp3} shows the comparison results. Other MEF methods can not deal with dynamic objects/texture properly so the tree leaves in the middle of their results suffer different degrees of fuzziness (Fig.~\ref{fig:comp3}(b)-Fig.~\ref{fig:comp3}(h)). Li~\emph{et al.}~\cite{Li2012Fast} and Li~\emph{et al.}~\cite{Li2013Image} treat RGB channels separately, making it difficult to make proper use of color information. As a result, the color in the wall regions appears dreary in Fig.~\ref{fig:comp3}(b) and Fig.~\ref{fig:comp3}(c). The method of~\cite{shen2014exposure} and~\cite{Shen2011Generalized} produce images with uncomfortable color (Fig.~\ref{fig:comp3}(d) and Fig.~\ref{fig:comp3}(e) appear pale outdoor scenes). Paul~\emph{et al.}~\cite{Paul2016Multi} reconstruct images using Haar wavelet which is easy to blur structural information such as the words in blackboard in Fig.~\ref{fig:comp3}(f). Mertens~\emph{et al.}'s method~\cite{Mertens2007Exposure} can not recover information of over-exposed regions. The outdoor scenes are too bright in Fig.~\ref{fig:comp3}(g). We apply Ma's method~\cite{ma2018multi} to improve the performance of Mertens~\emph{et al.}'s result. Ma~\emph{et al.} optimize existing MEF algorithms using a structural similarity index. As shown in Fig.~\ref{fig:comp3}(h), the method really works. However, blurry artifacts still exist in their result. Our method can maintain the color and structural information simultaneously and generate natural fusion results.

We implement Mertens~\emph{et al.}'s~\cite{Mertens2007Exposure} approach by ourselves rather than utilize the procedure which is integrated into OpenCV because the OpenCV implementation produces results that lost information. An example is shown in Fig.~\ref{fig:exposurefusion}. Fig.~\ref{fig:exposurefusion}(a) displays input images that are fully aligned by our method. Fig.~\ref{fig:exposurefusion}(b) shows the fusion result implemented by OpenCV which loses details in the window regions. Our implementation is more sensitive to such examples and generates better results. Moreover, we discovered that the whole tone of Fig.~\ref{fig:exposurefusion}(b) is not as natural as our result. Its upper regions are more bright and nether regions are more dark.


\section{Conclusion}

We have presented a hybrid synthesis method for multi-exposure fusion. The proposed method combines the strengths of flow-based methods and patch-based methods. Optical flow is applied for global registration which guarantees the computation efficiency. PatchMatch is used to further align error regions caused by optical flow, which can fully exclude moving objects. We also employ superpixel segmentation to improve the accuracy of detecting incorrect regions. The results of the proposed method own high-quality and natural color. We validate our method on various examples. The results are evaluated both qualitatively and quantitatively to demonstrate the effectiveness.


\ifCLASSOPTIONcaptionsoff
  \newpage
\fi



\end{document}